\documentclass[runningheads]{llncs}

\usepackage{eccvabbrv}

\usepackage{graphicx}
\usepackage{booktabs}

\usepackage[accsupp]{axessibility}  % Improves PDF readability for those with disabilities.
\usepackage{xcolor}         % colors
\usepackage{amsmath}
\usepackage{bm}
\usepackage{color, colortbl}
\definecolor{front-color}{HTML}{F1FFFF}
\definecolor{Gray}{gray}{0.90}
\usepackage{arydshln}
\usepackage{comment}
\usepackage{array}
\usepackage{multirow}
\usepackage{booktabs}
\usepackage[table]{xcolor} % Required for \rowcolor
\usepackage{graphicx}      % Required for \resizebox and \rotatebox
\usepackage{booktabs}

\usepackage{lipsum}

\usepackage{orcidlink}

\definecolor{mygray}{rgb}{0.89, 0.93, 0.85}
\definecolor{whitesmoke}{rgb}{0.96, 0.96, 0.96}
\definecolor{timberwolf}{rgb}{0.86, 0.84, 0.82}

\usepackage[capitalize]{cleveref}
\crefname{section}{Sec.}{Secs.}
\Crefname{section}{Section}{Sections}
\Crefname{table}{Table}{Tables}
\crefname{table}{Tab.}{Tabs.}
\usepackage{xurl}
\usepackage{array}
\usepackage{booktabs}

\usepackage{xspace}
\definecolor{light-gray}{gray}{0.6}
\definecolor{Gray}{gray}{0.9}
\definecolor{front-color}{HTML}{F5FFFA}

\newcommand{\ours}{Mobile-VideoGPT\xspace}

\begin{document}

\title{Mobile-VideoGPT: Fast and Accurate Model for Mobile Video Understanding}

\titlerunning{Mobile-VideoGPT}
\author{
Abdelrahman Shaker$^{1,*}$ \and
Muhammad Maaz$^{1}$ \and
Chenhui Gou$^{2}$ \and \\
Hamid Rezatofighi$^{2}$ \and
Salman Khan$^{1}$ \and
Fahad Shahbaz Khan$^{1,3}$
}
\authorrunning{A. Shaker et al.}
\institute{
$^{1}$Mohamed Bin Zayed University of Artificial Intelligence \quad \\
$^{2}$Monash University \quad
$^{3}$Linköping University\\[1mm]
{\small $^{*}$Corresponding author: \texttt{abdelrahman.youssief@mbzuai.ac.ae}}
}

\maketitle

\begin{abstract}
 Video understanding models often struggle with high computational requirements, extensive parameter counts, and slow inference speed, making them inefficient for practical use on edge devices. To tackle these challenges, we propose Mobile-VideoGPT, an efficient multimodal framework designed to operate with fewer than a billion parameters. Unlike traditional video large multimodal models (LMMs), Mobile-VideoGPT consists of lightweight dual visual encoders, efficient projectors, and a small language model (SLM), enabling real-time inference on resource-constrained platforms. To further improve efficiency, we present an Attention-Based Frame-Scoring mechanism to select the key-frames, along with an efficient token projector that prunes redundant visual tokens and preserves essential contextual cues. We evaluate our model across well-established six video understanding benchmarks (e.g., MVBench, EgoSchema, NextQA, and PerceptionTest), and our results demonstrate that Mobile-VideoGPT-0.5B achieves up to 7.3 tokens per second on NVIDIA Jetson Orin edge devices and 45.9 tokens per second on a single NVIDIA RTX A6000. Despite its compact size, it outperforms existing state-of-the-art 0.5B-parameter models by an average of 6 points, while using 40\% fewer parameters and delivering more than 2$\times$ higher throughput across both edge and desktop platforms. Our code and models are publicly available at: \href{https://github.com/Amshaker/Mobile-VideoGPT}{https://github.com/Amshaker/Mobile-VideoGPT}.
\end{abstract}    
\section{Introduction}
\label{sec:intro}

\begin{figure}[t]
  \centering
    \includegraphics[width=1.0\linewidth]{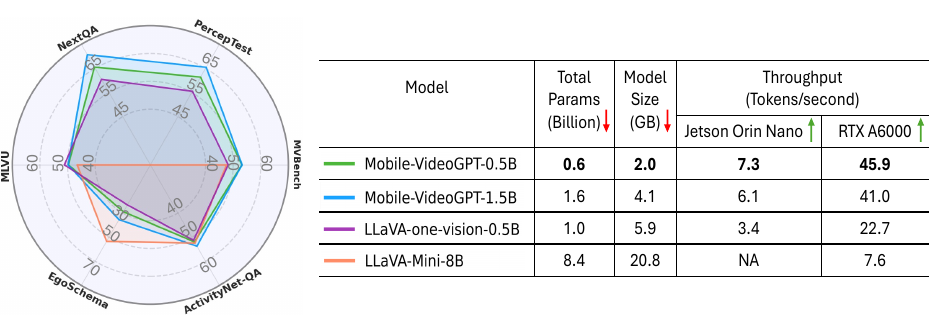}
    \vspace{-3mm}
    \caption{\textbf{Left}: Performance comparison of Mobile-VideoGPT against state-of-the-art efficient video understanding models across six benchmarks (PercepTest, NextQA, MVBench, EgoSchema, ActivityNet-QA, and MLVU). Mobile-VideoGPT achieves consistently strong accuracy across diverse video reasoning tasks. \textbf{Right}: Efficiency comparison in terms of total parameters, model size, and inference throughput (tokens/second) on Jetson Orin Nano and RTX A6000. Despite using significantly fewer parameters and a smaller memory footprint, Mobile-VideoGPT has higher throughput with better accuracy–efficiency trade-off and strong suitability for edge devices.}
    \label{fig:introduction}
\vspace{-1em}
\end{figure}

Multimodal Large Language Models (MLLMs) have emerged as a pivotal area of research and demonstrated exceptional abilities in video understanding tasks, including video captioning, visual question-answering, and visual grounding~\cite{video-llava,Maaz2023VideoChatGPT,li2024llava}. Remarkable performance on these tasks can be witnessed in proprietary models such as GPT-4o~\cite{openai2024gpt4o} and Gemini-1.5-Pro~\cite{geminiteam2024gemini15} or open-source alternatives such as Video-LLaVA~\cite{video-llava} and Video-ChatGPT~\cite{Maaz2023VideoChatGPT}. Despite the promising performance of these methods across specific benchmarks, deploying these models on edge devices presents a significant challenge. 

The design of large multimodal architectures often results in a substantial number of parameters and high computational demands, rendering them impractical for resource-constrained environments.  For instance, the recently introduced LLaVa-Mini~\cite{llavamini} proposes an image and video MLLM based on a single one‐vision token, yet its total number of parameters is 8.4B, and the model size is more than 20GB, making it impractical for real-time performance and edge deployment. Hence, there is a need for efficient video understanding models and training paradigms that are capable of retaining strong performance without exceeding the resource limitations.

Over the past decade, techniques such as knowledge distillation~\cite{hinton2015distilling,romero2014fitnets}, model pruning~\cite{han2015deepcompression,he2017channelpruning}, efficient attention~\cite{Shaker_2023_ICCV,Maaz2022EdgeNeXt}, and parameter-efficient fine-tuning~\cite{houlsby2019parameter} have partially addressed the challenge of reducing model complexity under resource constraints. However, these strategies come with notable performance trade-offs or require specialized hardware accelerators. Moreover, while they may effectively reduce the size of individual components (e.g., visual or text encoders), ensuring efficient projection and integration across modalities for video understanding and with advanced capabilities remains a challenge.

To this end, we introduce Mobile-VideoGPT, an efficient video understanding language model that balances between robust performance and high throughput on edge devices (i.e, Jetson Orin Nano~\cite{nvidia2022jetson}). We demonstrate the efficiency and the effectiveness of Mobile-VideoGPT across multiple video conversation benchmarks, including ActivityNet-QA, EgoSchema, MLVU, MVBench, NextQA, and PerceptionTest, where it not only is faster than the previous SoTA, but also outperforms them (See Fig.~\ref{fig:introduction}). Our contributions in this  work are as follows:

\begin{enumerate}
\setlength{\itemsep}{0pt}

\item A lightweight multimodal model designed specifically for real-time video understanding with a focus on edge device deployment.

\item We propose a novel frame scoring and selection strategy that optimally selects the representative video key-frames, along with efficient token projection, to enhance the model's ability to maintain visual fidelity while reducing computational overhead for efficient processing.

\item Our architecture is carefully optimized for edge deployment, featuring a compact model size of just 1GB in FP16 precision and requiring only 3GB of VRAM. Owing to its efficient design, it achieves a high throughput of 7.3 tokens per second on Jetson Orin Nano~\cite{nvidia2022jetson}.

\item Mobile-VideoGPT showcases superior performance compared to competitive models across six video understanding benchmarks, achieving an average improvement of 6 points over LLaVA-OneVision-0.5B~\cite{li2024llava}. Our model is also 2$\times$ faster, while being less than 3$\times$ its size. \end{enumerate}
%%%%%%%%%%%%%%%%%%%%%%%%%%%%% Related Work Section %%%%%%%%%%%%%%%%%%%%%%%%%%%%% 
\section{Related Work}
\label{sec:related}

\noindent{\textbf{Small Language Models (SLMs)}}:
In the past year, the rise of large language models (LLMs) has driven significant progress in natural language processing, including proprietary models like GPT-4~\cite{2023GPT4VisionSC}, and open-source models such as LLaMA~\cite{touvron2023llama1} and DeepSeek~\cite{liu2024deepseek}. Although these models have sparked increased interest in enhancing the performance of LLMs, the large size of these models often leads to substantial memory consumption and slow processing speeds. Moreover, the energy demands associated with training and operating these models raise important sustainability issues, which are becoming increasingly urgent as LLMs expand in scale and complexity. To address this, there has been a growing trend in the development of small and medium-sized LLMs, such as Qwen2-0.5B~\cite{yang2024qwen2} and Llama3.2-1B~\cite{llama3}, specifically designed for resource-constrained environments. Given the limited computational capabilities of edge devices such as mobile phones, autonomous vehicles, and embodied AI systems, several lightweight LLMs~\cite{li2023textbooks,zhang2024tinyllama,liu2024mobilellm,chu2023mobilevlm} have gained significant attention. TinyLLaMA~\cite{zhang2024tinyllama} conducts great open-source work on their 1.1B base and chat models with the same architecture and tokenizer as LLaMa2~\cite{touvron2023llama2}. MobileLLaMA~\cite{chu2023mobilevlm} scales down the architecture of LLaMA and releases 1B and 3B models trained from scratch with public datasets. Recently, MobileLLM~\cite{liu2024mobilellm} optimizes sub-billion parameter models by proposing an immediate block-wise weight-sharing approach. These compact models are optimized to deliver efficient performance while addressing the unique constraints of resource-limited environments. Another vital line of research involves applying model compression techniques for LLMs, such as pruning \cite{sparsegpt}, quantization \cite{xiao2023smoothquant,li2024norm}, and low-rank decomposition \cite{yao2024exploring}. These techniques reduce the computational demands of LLMs and enhance their practicality for deployment on resource-constrained devices.

\noindent{\textbf{Vision Language Models}}: The advancement of visual language models (VLMs) has been significantly shaped by the introduction of CLIP~\cite{clip}, which utilizes a contrastive loss to align vision and language embeddings within a unified representation space. Recent progress in LLMs has further enhanced the development of VLMs, as demonstrated in Kosmos-2~\cite{kosmos-2}, GLaMM~\cite{hanoona2023GLaMM}, BLIP-2~\cite{li2023blip}, MiniGPT-4~\cite{zhu2023minigpt}, LLaVA~\cite{liu2023llava} and PALO~\cite{PALO}. These models combine visual and linguistic features, leveraging advanced cross-modal designs, such as cross-attention mechanisms, Q-Former modules, or simple multi-layer perceptron layers.

\noindent{\textbf{Video Language Models}}: Current video understanding models typically rely on image encoders~\cite{2023videochat,goldfish,video-llava}, video encoders~\cite{Maaz2024VideoGPT+,Li_2024_VideoMamba}, or dual spatial/temporal vision stacks~\cite{chung2025unifying,damonlpsg2024videollama2} to process visual inputs alongside advanced LLMs. Image encoders capture rich spatial details from frame sequences but often miss explicit temporal cues essential for complex action recognition. Video encoders provide temporal context but are constrained by efficiency, typically processing sparse frames at lower resolutions and thus losing both contextual and spatial fidelity. VideoMamba~\cite{li2024videomamba}, a recent parameter-efficient video encoder, addresses these issues by extending the Mamba architecture to the video domain, combining linear-complexity operators for efficient long-term modeling with strong scalability.  

Goldfish~\cite{goldfish} and VideoGPT+~\cite{Maaz2024VideoGPT+} improve scalability by segmenting videos into short clips processed independently. LLAVA-OneVision~\cite{li2024llava} advances this direction by introducing a unified framework for both image and video tasks, using a single vision encoder for frame-level feature extraction while relying on the language model’s positional embeddings for temporal understanding. Despite these advances, most approaches remain impractical for edge deployment due to high computational demands, large memory footprints, and reliance on powerful accelerators. Their complex architectures with large encoders and high-resolution inputs require substantial GPU and storage resources, limiting their use in real-time, resource-constrained environments.  

More recently, attention-style frame selection~\cite{liang2024keyvideollmlargescalevideokeyframe,TangQiuXieTianJiaoYe_2025_AdaptiveKeyframeSampling,Buch2025_FlexibleFrameSelection} and efficient projection layers to reduce visual tokens~\cite{TokenPacker,wang2024llavazipadaptivevisualtoken,huang2024prunevid} have emerged to mitigate these challenges. These limitations also motivate our work to develop a lightweight, resource-efficient video understanding model dedicated to edge devices by reducing the model size and memory usage and enabling real-time inference without compromising performance.

%%%%%%%%%%%%%%%%%%%%%%%%%%%%% Method Section %%%%%%%%%%%%%%%%%%%%%%%%%%%%% 
\section{\ours}
\label{sec:method}
We introduce Mobile‐VideoGPT, specifically designed for efficient video understanding. It incorporates both an image encoder and a video encoder; each optimized for speed and number of parameters, along with compact, efficient projection modules that project the latent visual features into a suitable format for small language model.

\begin{figure*}[t]
  \centering
    \includegraphics[width=.97\linewidth]{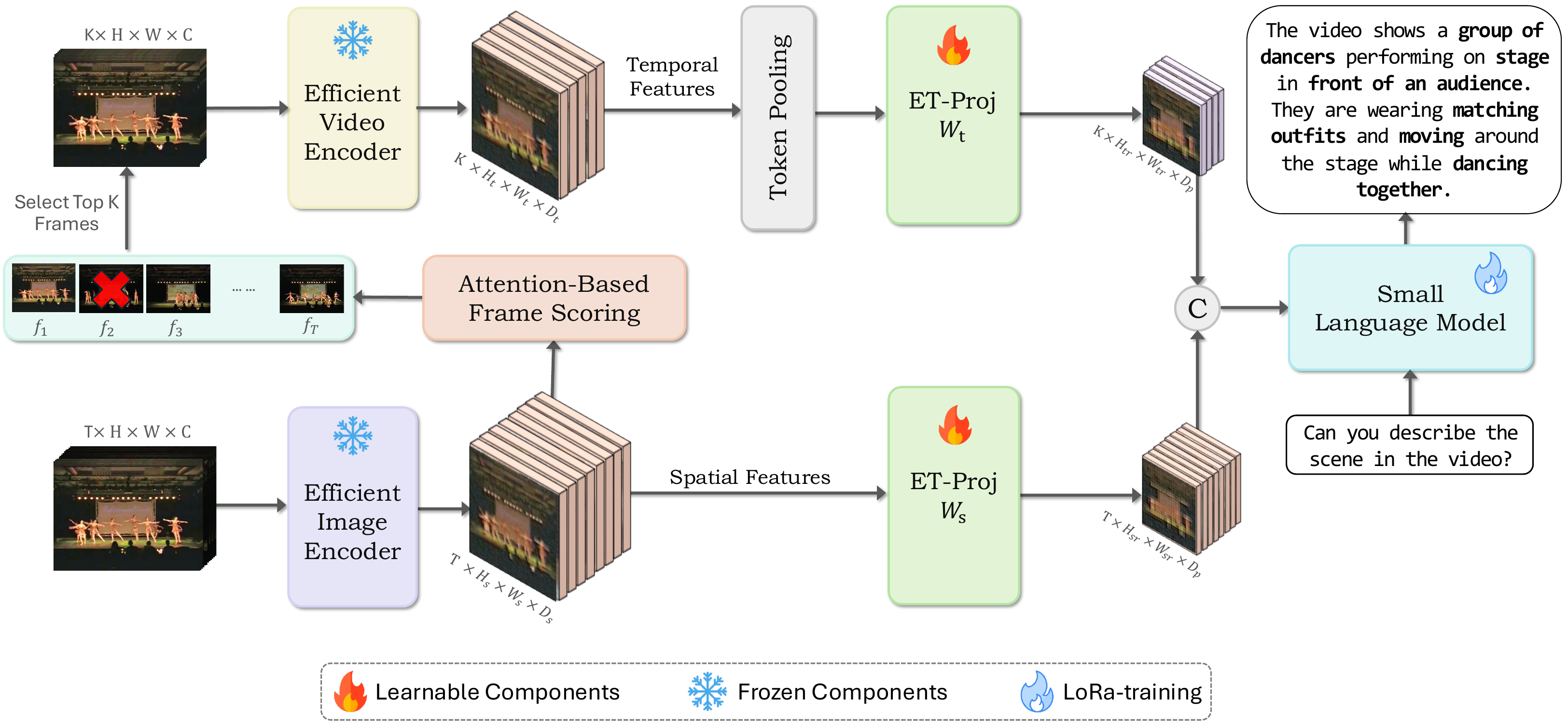}
    
    \caption{\small \textbf{Overview of the proposed Mobile-VideoGPT}. The pipeline begins by extracting spatial features from all video frames with an efficient image encoder. These features then flow into an attention-based frame scoring mechanism, which identifies the top salient K frames. Next, an efficient video encoder processes these selected frames to capture temporal dynamics. The resulting spatial and temporal representations are projected into a unified vision–language space using an efficient token projector (ET‐Proj). Finally, the projected tokens are fused, and a small language model leverages these tokens to generate comprehensive responses to video‐based questions.}
    \label{fig:method_overall}
\end{figure*}

\subsection{Overall Architectural Design}

\noindent
Inspired by recent work~\cite{Maaz2024VideoGPT+,zohar2024apollo}, {Mobile‐VideoGPT} adopts a dual‐encoder design combined with a small language model (SLM). As illustrated in Fig.~\ref{fig:method_overall}, the proposed framework consists of: (1)~a CLIP‐based image encoder~\cite{clip} for capturing spatial and semantic features, (2)~a lightweight VideoMamba encoder~\cite{li2024videomamba} for extracting temporal information, (3)~an Efficient Token Projection ($ET_{\text{Proj}}$) module for feature reduction and projection, and (4)~an SLM for generating context‐rich textual responses.

\noindent
We start by uniformly sampling $T$ frames from the input video. For each frame $t \in \{1,\dots,T\}$, we extract spatial embeddings via the CLIP‐based image encoder, forming
\[
  \mathbf{F}_{\text{clip}} \;\in\; \mathbb{R}^{\,T \,\times\, H_s \,\times\, W_s \,\times\, D_s},
\]
where $H_s$ and $W_s$ denote the spatial dimensions for image features, and $D_s$ the spatial feature depth. Then, we apply attention‐based frame scoring (Section~\ref{sec:frame_scoring}) to select the top‐$K$ key-frames, which are passed to the VideoMamba encoder-M to encode the temporal features. This yields
\[
  \mathbf{F}_{\text{vmamba}} \;\in\; \mathbb{R}^{\,K \,\times\, H_t \,\times\, W_t \,\times\, D_t},
\]
where $H_t$ and $W_t$ represent spatial dimensions for the video features, and $D_t$ is the temporal feature depth.

\noindent
Next, both the clip‐based and the VideoMamba embeddings are processed through Efficient Token Projection modules. Concretely, $ET_{\text{Proj}}$ reduces and transforms each modality into a unified vision--language embedding space:
\begin{align}
  \widehat{\mathbf{F}}_{\text{clip}} 
    &= \mathrm{ET_{\text{Proj}}}_s \bigl(\mathbf{F}_{\text{clip}}\bigr)
     \;=\; \mathbf{F}_{\text{clip}} \,\mathbf{W}_s, 
   \label{eq:clipprojection} \\
  \widehat{\mathbf{F}}_{\text{vmamba}} 
    &= \mathrm{ET_{\text{Proj}}}_t \bigl(\mathbf{F}_{\text{VMamba}}\bigr)
     \;=\; \mathbf{F}_{\text{VMamba}} \,\mathbf{W}_t, 
   \label{eq:videoprojection}
\end{align}
where $\mathbf{W}_s$ and $\mathbf{W}_t$ are learnable weight matrices for spatial and temporal features, respectively. We then fuse $\widehat{\mathbf{F}}_{\text{clip}}$ and $\widehat{\mathbf{F}}_{\text{vmamba}}$ into a single token sequence, which a \emph{small language model (SLM)} consumes to produce context‐rich responses to video‐based queries.

\subsection{Frame Selection \label{frameselect}}
\label{sec:frame_scoring}

A straightforward approach would be to feed all \(T\) sampled frames directly into the video encoder. However, this often proves redundant and inefficient. Therefore, we propose an \emph{attention‐based frame scoring} mechanism to select only the most salient \(K\) frames (\(K < T\)), substantially reducing the computational load. In our setup, \(K = T/2\). Let \(\mathbf{F}_{\text{clip}} \in \mathbb{R}^{T \times H_s \times W_s \times D_s}\) be the image features from the CLIP‐based encoder. We flatten all \(T\) frames into \(\mathbf{F}_{\text{sp}} \in \mathbb{R}^{S \times D_f}\), where \(S = T \times H_I \times W_I\) and \(D_f\) is the spatial feature dimension. We then compute a \emph{spatial attention} matrix:
\begin{equation}\label{equ:SA}
    \text{SA} \;=\; \mathrm{Softmax}\!\Bigl[\tfrac{\mathbf{F}_{\text{sp}}\,\mathbf{F}_{\text{sp}}^{\!\top}}{\sqrt{d_f}}\Bigr],
\end{equation}
where \(\mathbf{F}_{\text{sp}}\) are the flattened spatial features. Next, we sum the attention scores each token receives to produce a frame‐wise importance vector \(T_{\text{score}} \in \mathbb{R}^T\). We then select the top‐\(K\) frames via:
\begin{equation}\label{equ:topK}
    \text{KF} \;=\; \mathrm{Top\mbox{-}K}\bigl(T_{\text{score}},\,K\bigr).
\end{equation}

\noindent
Finally, we feed the selected key-frames (\text{KF}), after reshaping, into {VideoMamba} to capture temporal dynamics, yielding \(\mathbf{F}_{\text{vmamba}} \in \mathbb{R}^{K \times H_t \times W_t \times D_t}\). In our experiments, we set \(T=16\) and \(K=8\). By discarding low‐importance frames, this approach effectively reduces redundant computations while preserving crucial contextual information.

\subsection{Efficient Token Projection}
\label{sec:etp}

\begin{figure*}[t]
  \centering
    \includegraphics[width=.95\linewidth]{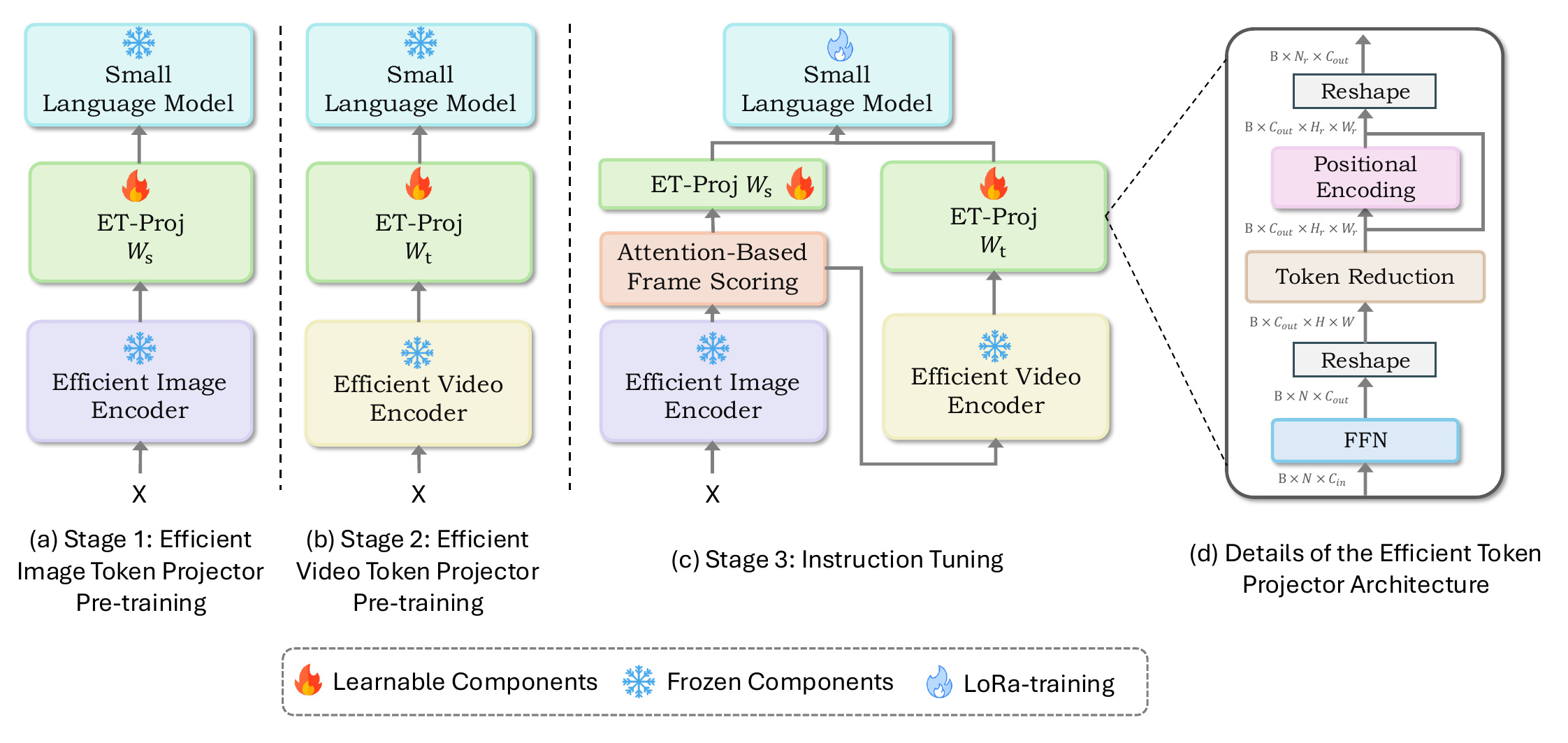}
    
    \caption{\textbf{Overview of Mobile-VideoGPT training strategy and the architecture of the Efficient Token Projector (ET-Proj).} (a–c) Training proceeds in three stages: pre-training the image token projector with an efficient image encoder, extending to video projector pre-training with a video encoder, and finally instruction tuning where both projectors are learnable and LoRA fine-tuning is applied to the language model. (d) ET-Proj uses a lightweight feedforward network (FFN) to refine token embeddings, a token reduction step via global average pooling, and a positional encoding module with skip connection to retain spatial and temporal context.}

    \label{fig:method_overall_2}
    
\end{figure*}

Fig.~\ref{fig:method_overall_2} (d) describes the \emph{Efficient Token Projector} ($ET_{Proj}$), 
which compresses and re‐encodes tokenized image/video features while preserving crucial 
spatiotemporal information. The module comprises three stages: (1)~Feed‐forward network 
(FFN) that transforms the input features 
\(\mathbf{x} \in \mathbb{R}^{B \times N \times C_{\mathrm{in}}}\) 
into 
\(\mathbb{R}^{B \times N \times C_{\mathrm{out}}}\), where $B$ is the batch size, $N$ is the visual tokens, and $C$ is the hidden dimension. Then, we reshape the projected tensor into 
\(\mathbf{X} \in \mathbb{R}^{B \times C_{\mathrm{out}} \times H \times W}\).
(2)~Token reduction with \(\mathrm{AdaptivePool}\) to reduce the spatial size \((H,\,W)\) to a smaller 
\((H_r,\,W_r)\), and (3)~Convolution‐based positional encoder with skip connection to enhance the key positional 
dependencies via learned offsets. Formally, the block can be summarized as:
\begin{equation}\label{eq:etproj}
   \mathbf{X}'' 
   = \mathrm{PosEnc}\Bigl(\mathrm{TokenReduction}\bigl(\mathrm{FFN}(\mathbf{x})\bigr)\Bigr).
\end{equation}
By reducing the spatial tokens while retaining global features and positional cues, our projector provides an efficient mechanism for compressing high‐dimensional tokens in both images and videos, 
substantially reducing computational overhead while preserving spatiotemporal fidelity.

%%%%%%%%%%%%%%%%%%%%%%%%%%%%% Experiments %%%%%%%%%%%%%%%%%%%%%%%%%%%%% 
\section{Experiments}
\label{sec:exp}
\subsection{Evaluation Benchmarks}
We conduct a quantitative evaluation of Mobile-VideoGPT across \textit{\textbf{six well-established benchmarks}}, namely: i) ActivityNet-QA~\cite{yu2019activitynet}, ii) EgoSchema~\cite{mangalam2023egoschema}, iii) MLVU~\cite{zhou2024mlvu}, iv) MVBench~\cite{li2023mvbench}, v) NExTQA~\cite{xiao2021next}, and vi) PerceptionTest~\cite{puatruaucean2023perception}.
These benchmarks collectively cover diverse tasks such as multiple‐choice question answering, open‐ended generation, spatiotemporal reasoning, and detailed action understanding, offering a comprehensive evaluation of our model's capabilities across real‐world and research‐oriented scenarios.
Specifically, \textbf{ActivityNet-QA}~\cite{yu2019activitynet} evaluates activity and action understanding on a test set of 8K samples derived from the ActivityNet dataset, using GPT-3.5-Turbo for open-ended evaluation following~\cite{Maaz2023VideoChatGPT}.
\textbf{EgoSchema}~\cite{mangalam2023egoschema} targets egocentric video understanding through 5K MCQ samples sourced from Ego4D~\cite{grauman2022ego4d}, with results evaluated via an online server.
\textbf{MLVU}~\cite{zhou2024mlvu} probes long video comprehension across nine distinct tasks spanning both MCQ and open-ended generation, and we report results on the 2,593-sample dev set following~\cite{li2024llava}.
\textbf{MVBench}~\cite{li2023mvbench} offers a broad assessment of video understanding in the wild through 20 fine-grained sub-tasks, ranging from action sequence and prediction to egocentric navigation and counterfactual inference—totaling 4K MCQs.
\textbf{NExTQA}~\cite{xiao2021next} focuses on temporal action understanding, evaluating causal and temporal action reasoning as well as scene comprehension on an MCQ test set of 8,564 samples.
Finally, \textbf{PerceptionTest}~\cite{puatruaucean2023perception} assesses perception and reasoning capabilities on a validation set of 19,140 samples, ensuring a thorough evaluation of our model's real-world applicability.
More details are provided in the supplementary material.

%%%%%%%%%%%%%%%%%%%%%%%%%%%%% Implementation Details %%%%%%%%%%%%%%%%%%%%%%%%%%%%% 
\subsection{Implementation Details}

In our experiments, we utilize CLIP-B/16~\cite{clip} as the image encoder, and for the video encoder, we use VideoMamba-M~\cite{li2024videomamba}, alongside the Qwen-2.5, 0.5/1.5B~\cite{yang2024qwen2} LLM. Both image and video encoders process inputs at a resolution of $224\times224$. Our training pipeline consists of two warm-up phases followed by an instruction-tuning phase. During warm-up, we separately train with the image and video encoders on the CC-595K dataset~\cite{liu2023improved}, optimizing only the vision-to-language adapters while keeping the remaining model parameters frozen. In the instruction-tuning phase, we apply LoRA~\cite{hu2022lora} with a rank of $r=64$ to the LLM, while the vision-to-language adapters are fully trained and the vision encoders remain frozen. The learning rate is set to $1e^{-3}$ during warm-up and $2e^{-4}$ during instruction tuning. During instruction tuning, we first sample videos at 1 FPS, followed by a uniform sampling of 16 frames. The value of K is set to 8 for attention-based frame scoring.

We use instruction-tuning data from VideoGPT+~\cite{Maaz2024VideoGPT+}, which consists of Kinetics-710~\cite{kay2017kinetics}, Something-Something-v2~\cite{goyal2017something}, conversations from VideoChat~\cite{2023videochat}, CLEVRER~\cite{yi2019clevrer}, the VQA dataset from WebVid~\cite{bain2021frozen}, and NExT-QA~\cite{xiao2021next}, along with the PerceptionTest~\cite{puatruaucean2023perception} subset from LLaVA-178K~\cite{zhang2024videoinstructiontuningsynthetic} and a subset of VideoChat2-IT~\cite{li2023mvbench}. The instruction-tuning dataset amounts to approx. 1M samples, and we train our model for two epochs. 

Although we use only a fraction of the instruction-tuning dataset compared to previous SoTA approaches (e.g., LLaVA-OneVision, which utilizes approx.~4.8M samples combining images, multi-image, and videos), our ablations indicate that the observed performance gains are primarily driven by our architectural improvements—specifically, the dual-encoder design, frame-selection strategy, and efficient projection modules. We anticipate that scaling to a larger instruction-tuning dataset would further improve performance, but the current results demonstrate that our efficient MLLM architecture can achieve better results even with less number of samples.

\begin{table}[t]
\centering
\caption{Performance comparison of Mobile-VideoGPT against state-of-the-art video understanding models across six benchmarks. We report accuracy (\%) alongside efficiency metrics including total parameters, input resolution, frame count, and throughput. Our models achieve competitive performance with significantly higher throughput (frames per second) than larger counterparts. Throughput is evaluated on a single NVIDIA RTX A6000 GPU.}
\label{tab:video-bench}
\small
\setlength{\tabcolsep}{3.5pt} 
\renewcommand{\arraystretch}{1.1}
\resizebox{\columnwidth}{!}{%
\begin{tabular}{l cccc cccccc}
    \toprule
    \multirow{2}{*}{\textbf{Model}}
    & \rotatebox{90}{\textbf{Total Params}}
    & \rotatebox{90}{\textbf{Input Res.}}
    & \rotatebox{90}{\textbf{Num Frames}}
    & \rotatebox{90}{\textbf{Throughput}}
    & \rotatebox{90}{\textbf{\footnotesize{ActNet-QA}}}
    & \rotatebox{90}{\textbf{\footnotesize{EgoSchema}}}
    & \rotatebox{90}{\textbf{\footnotesize{MLVU}}}
    & \rotatebox{90}{\textbf{\footnotesize{MVBench}}}
    & \rotatebox{90}{\textbf{\footnotesize{NextQA}}}
    & \rotatebox{90}{\textbf{\footnotesize{PercepTest}}} \\ 
    \cmidrule(lr){2-5} \cmidrule(lr){6-11}
     & $\downarrow$ & $\downarrow$ & $\downarrow$ & $\uparrow$ & test & test & m-avg & test & mc & val \\ 
    \midrule
    
    \rowcolor{gray!10}
    GPT-4V~\cite{openai2023gpt4v} & - & 512$\times$512 & 32 & - & 57.0 & - & 49.2 & 43.5 & - & - \\
    \rowcolor{gray!10}
    Gemini-1.5-Flash~\cite{team2023gemini} & - & 768$\times$768 & 32 & - & 55.3 & 65.7 & - & - & - & - \\
    \rowcolor{gray!10}
    Gemini-1.5-Pro~\cite{team2023gemini} & - & 768$\times$768 & 32 & - & 57.5 & 72.2 & - & - & - & - \\
    \midrule 
    \multicolumn{11}{c}{\textit{$>$7B Models}} \\ \midrule

    LLaMA-VID-7B~\cite{li2024llamavid} & 7.3B & 336$\times$336 & 1fps & 6.0 &  47.4 & 38.5 & 33.2  & 41.4 & - & - \\
    Video-LLaVA-7B~\cite{lin2023video} & 7.3B & 224$\times$224 & 8 & 12.5 & 38.4 & 45.3 & - & 43.1 & - & -\\
    VideoChat2-7B~\cite{2023videochat} & 7.3B & 224$\times$224 & 16 & 11.4 & 49.1 & 54.4 & - & 60.4 & 79.5 & 47.3 \\
    LongVA-7B~\cite{zhang2024long} & 7.4B & 224$\times$224 & 8 & 9.2 & 50.0 & - & 56.3 & - & 68.3 & - \\
    LLaVA-Mini-8B~\cite{llavamini} & 8.4B & 336$\times$336 & 1fps & 4.6 & 52.3 & 51.2 & 42.8 & 44.5 & - & - \\ \midrule 

    \multicolumn{11}{c}{\textit{$<$7B Models}} \\ \midrule
    InternVL2.5-2B~\cite{chen2024expanding} & 2.2B & 448$\times$448 & 32 & 21.8 & 54.1 & - & 51.6 & 57.5 & 75.6 & 66.3 \\
    Video-ChatGPT-3.8B~\cite{Maaz2023VideoChatGPT} & 4.2B & 224$\times$224 & 100 & 14.1 & 35.2 & 36.2 & 31.3 & 32.7  & - & - \\
    LLaVA-OneVision-0.5B~\cite{li2024llava} & 1.0B & 384$\times$384 & 32 & 22.7 & 50.5 & 26.8 & 50.3 & 47.2 & 57.2 & 49.2 \\

    \rowcolor{front-color}
    \textbf{Mobile-VideoGPT-0.5B} & \textbf{0.6B} & \textbf{224$\times$224} & \textbf{16} & \textbf{45.9} & \textbf{51.6} & \textbf{31.4} & \textbf{47.9} & \textbf{53.5} & \textbf{65.4} & \textbf{58.7} \\     
    \rowcolor{front-color}
    \textbf{Mobile-VideoGPT-1.5B} & \textbf{1.6B} & \textbf{224$\times$224} & \textbf{16} & \textbf{41.0} & \textbf{54.4} & \textbf{36.7} & \textbf{48.1} & \textbf{53.6} & \textbf{73.7} & \textbf{65.3} \\     
    \bottomrule
\end{tabular}%
}
\end{table}

\subsection{Quantitative Results}
We provide a detailed comparison of our models in terms of speed and accuracy with recent methods in Tab.~\ref{tab:video-bench}. For efficiency, we compare total parameters, input resolution, number of input frames, and throughput. Overall, our Mobile-VideoGPT-0.5B variant performs comparably to LLaVA-OneVision~\cite{li2024llava} while being twice as fast (45.9 vs. 22.7 tokens/sec for LLaVA-OneVision-0.5B). Similarly, our 1.5B variant achieves results close to the recently released LLaVA-Mini-8B~\cite{llavamini} while running up to 9$\times$ faster (41.0 vs. 4.6 tokens/sec for LLaVA-Mini-8B). These results demonstrate the advantages of our architectural design and training approach, enabling strong performance while maintaining efficiency. On~ActivityNet-QA, our 0.5B model achieves 51.6\% accuracy, outperforming LLaVA-OneVision-0.5B by 1.1\%. Our 1.5B model achieves a score of 54.4\%, surpassing LLaVA-Mini-8B by 2.1\% and InternVL-2.5-2B by 0.3\%, and coming just 0.9\% behind the closed-source Gemini-1.5-Flash. Since ActivityNet-QA focuses on activity understanding in videos, the presence of the video encoder enables our model to capture rich temporal dependencies, leading to strong performance on activity-centric benchmarks. Although our model is not explicitly designed for long videos, our models still achieve competitive results on the long-video MLVU benchmark, highlighting the effectiveness of our dual-encoder design. Notably, the recently introduced InternVL2.5-2B achieves marginally higher average performance than Mobile-VideoGPT-1.5B. However, our model uses 30\% fewer parameters with 2$\times$ higher throughput.

Similarly, on MVBench, our 0.5B model achieves a score of 53.5, surpassing LLaVA-Mini-8B and LLaVA-OneVision-0.5B by 9\% and 6.3\%, respectively. MVBench evaluates video MLLMs across 20 temporal understanding sub-tasks, and we provide a detailed comparison of our 0.5B model against previous approaches for each sub-task in Tab.~\ref{tab:detail_mvbench}. Notably, our model performs well in tasks requiring strong temporal reasoning. For example, in the Moving Direction category, our 0.5B model scores 59.0\%, significantly higher than the second-best score of 31.0\% by LLaVA-Mini. Similar trends are observed in Action Localization, Action Count, Moving Count, Moving Attribute, and State Change sub-tasks. We attribute this strong performance to our attention-based frame selection algorithm coupled with the video encoder, which effectively captures video-specific dependencies. Notably, our model performs less effectively on fine-grained tasks, likely due to lower input resolution and the limited number of frames we use to maintain the efficiency of our design with real-time speed. A detailed discussion about the efficiency and performance on fine-grained tasks is provided in the Sec.~\ref{sec:ablations}

To evaluate open-ended generation, we use the VideoChatGPT benchmark~\cite{Maaz2023VideoChatGPT}. Since the original setup relied on GPT-3.5-turbo-0613 (which is now deprecated), we re-evaluated both LLaVA-OneVision-0.5B and Mobile-VideoGPT-0.5B with GPT-3.5-turbo-1106, achieving average scores of 2.60 and 2.68, respectively, demonstrating that our lightweight model attains superior open-ended performance while being significantly more efficient.

\begin{table}[t]
\centering
\caption{\textbf{Detailed performance on MVBench.} Comparison across 20 sub-tasks grouped into nine categories against Video-ChatGPT, Video-LLaMA, Video-LLaVA, LLaVA-Mini, and LLaVA-OneVision. Best results are in \textbf{bold} and second-best are \underline{underlined}. Despite its compact 0.5B scale, Mobile-VideoGPT achieves the highest overall average, and demonstrates particularly strong performance in counting, positional reasoning, attribute understanding, and object-centric tasks.}

\label{tab:detail_mvbench}
\scriptsize
\setlength{\tabcolsep}{2.5pt} % Tighten spacing for single column
\renewcommand{\arraystretch}{1.1}
\resizebox{\columnwidth}{!}{%
\begin{tabular}{ll cccccc}
\toprule
\textbf{Spatial} & \textbf{Temporal} & \textbf{V-ChatGPT} & \textbf{V-LLaMA} & \textbf{V-LLaVA} & \textbf{LLaVA-M} & \textbf{LLaVA-OV} & \textbf{Ours} \\ 
\cmidrule(lr){1-2} \cmidrule(lr){3-8}
\multicolumn{2}{l}{\textbf{Average}} & 32.7 & 34.1 & 43.1 & 44.5 & \underline{47.2} & \textbf{53.5} \\ \midrule

\multirow{5}{*}{\textbf{Action}} 
& Action Sequence     & 23.5 & 27.5 & 44.5 & 44.5 & \textbf{57.0} & \underline{54.0} \\
& Action Prediction   & 26.0 & 25.5 & 50.0 & 44.5 & \textbf{64.5} & \underline{56.5} \\
& Action Antonym      & \underline{62.0} & 51.0 & 49.0 & \textbf{76.0} & 49.5 & 56.0 \\
& Fine-grained Act.   & 22.5 & 29.0 & \textbf{42.0} & 37.0 & \underline{38.0} & 36.5 \\
& Unexpected Act.     & 26.5 & 39.0 & 54.5 & 58.5 & \textbf{70.0} & \underline{60.5} \\ \midrule

\multirow{3}{*}{\textbf{Object}} 
& Object Existence    & 54.0 & 48.0 & 52.5 & 50.0 & \underline{53.5} & \textbf{82.5} \\
& Object Interaction  & 28.0 & 40.5 & 46.5 & 50.0 & \textbf{68.5} & \underline{51.0} \\
& Object Shuffle      & \underline{40.0} & 38.0 & \textbf{40.5} & 29.5 & 37.5 & 37.5 \\ \midrule

\multirow{2}{*}{\textbf{Position}}  
& Moving Direction    & 23.0 & 22.5 & 27.0 & \underline{31.0} & 20.0 & \textbf{59.0} \\
& Action Localiz.     & 20.0 & 22.5 & 28.5 & \underline{32.5} & 27.0 & \textbf{38.5} \\ \midrule

\textbf{Scene} & Scene Trans.  
& 31.0 & 43.0 & 84.5 & \underline{85.5} & \textbf{87.0} & 82.0 \\ \midrule

\multirow{2}{*}{\textbf{Count}}
& Action Count        & 30.5 & 34.0 & \underline{44.5} & 35.0 & 40.5 & \textbf{52.5} \\
& Moving Count        & 25.5 & 22.5 & 26.5 & \underline{40.0} & 36.5 & \textbf{63.5} \\ \midrule

\textbf{Attribute} & Moving Attr.
& 39.5 & 32.5 & 53.0 & 48.0 & \underline{54.5} & \textbf{81.0} \\ \midrule

\multirow{2}{*}{\textbf{Pose}}
& State Change        & \underline{48.5} & 45.5 & 38.5 & 41.0 & 40.0 & \textbf{59.0} \\
& Fine-grained Pose   & 29.0 & 32.5 & \underline{34.0} & 29.5 & \textbf{43.0} & 28.0 \\ \midrule

\textbf{Character} & Char. Order    
& 33.0 & 40.0 & 42.5 & \underline{52.0} & \textbf{55.5} & 47.0 \\ \midrule

\multirow{3}{*}{\textbf{Cognition}}
& Ego. Navigation     & 29.5 & 30.0 & \textbf{32.5} & 31.0 & 26.5 & \underline{31.5} \\
& Episodic Reas.      & 26.0 & 21.0 & \underline{38.0} & \underline{38.0} & \textbf{38.5} & 37.0 \\ 
& Counterf. Inf.      & 35.5 & \underline{37.0} & 32.0 & 36.0 & 35.5 & \textbf{57.5} \\ \bottomrule    
\end{tabular}%
}
\end{table}

%%%%%%%%%%%%%%%%%%%%%%%%%%%%% Vision Encoder Type, Pooling Strategy & Dataset
\begin{table}[t!]
\centering
\caption{Systematic ablation study of architectural components. We evaluate the impact of vision encoder types, frame selection (FS) strategies, and token projection methods on performance and throughput. Highlights in \colorbox{front-color}{color} indicate our final design choices.}
\label{tab:encoder_pooling_and_data}
\scriptsize
\setlength{\tabcolsep}{2.8pt}
\renewcommand{\arraystretch}{1.2}
\resizebox{\columnwidth}{!}{
\begin{tabular}{l ccc cc cc}
\toprule
\multirow{2}{*}{\textbf{Benchmark}} & \multicolumn{3}{c}{\textbf{Vision Encoder}} & \multicolumn{2}{c}{\textbf{Frame Selection (FS)}} & \multicolumn{2}{c}{\textbf{Token Projection}} \\ 
\cmidrule(lr){2-4} \cmidrule(lr){5-6} \cmidrule(lr){7-8}
                  & \textbf{Image} & \textbf{Video} & \textbf{Dual w/ FS} & \textbf{w/o FS} & \textbf{Attn-Based} & \textbf{MLP$_{proj}$} & \textbf{ET$_{proj}$} \\ 
\midrule
ActNet-QA       & 51.4     & 49.3           & \cellcolor{front-color}\textbf{51.6}                       & 51.3        & \cellcolor{front-color}\textbf{51.6}         & 50.9       & \cellcolor{front-color}\textbf{51.6}          \\
EgoSchema       & 26.5           & 27.1           & \cellcolor{front-color}\textbf{31.4}                  & 31.2     & \cellcolor{front-color}\textbf{31.4}         & 30.5      & \cellcolor{front-color}\textbf{31.4}          \\
MLVU            & 43.0       & \textbf{48.2}        & \cellcolor{front-color}47.9          & \textbf{48.3}             & \cellcolor{front-color}{47.9}         & 45.0         & \cellcolor{front-color}\textbf{47.9}          \\
MVBench         & 52.4        & 49.6        & \cellcolor{front-color}\textbf{53.5}          & \textbf{54.0}               & \cellcolor{front-color}{53.5}         & \textbf{54.9}       & \cellcolor{front-color} 53.5      \\
NextQA          & 62.1        & 62.6       & \cellcolor{front-color}\textbf{65.4}          & 65.3               & \cellcolor{front-color}\textbf{65.4}         & 64.9         & \cellcolor{front-color}\textbf{65.4}          \\
PercepTest      & 58.3       & 58.0      & \cellcolor{front-color}\textbf{58.7}          & \textbf{58.7}        & \cellcolor{front-color}\textbf{58.7}         & \textbf{59.5}       & \cellcolor{front-color}58.7          \\ 
\midrule 
\textbf{Avg. Accuracy}        & 49.0        & 49.1           & \cellcolor{front-color}\textbf{51.4}          & \textbf{51.5}             & \cellcolor{front-color}{51.4}         & 51.0        & \cellcolor{front-color}\textbf{51.4}          \\ \midrule 

\textbf{Throughput}        & 49.6        & 41.3       & \cellcolor{front-color}\textbf{45.9}          & 38.5             & \cellcolor{front-color}\textbf{45.9}         & 39.2            & \cellcolor{front-color}\textbf{45.9}          \\ 
\bottomrule
\end{tabular}
}
\end{table}

\subsection{Ablations}
\label{sec:ablations}
% In Tab.~\ref{tab:encoder_pooling_and_data}, we provide ablation studies on our dual-encoder design in Mobile-VideoGPT, our frame-selection algorithm, and our token projection module. Our dual-encoder design performs better than the single-encoder variants (i.e., using either an image or a video encoder) across all benchmarks, except for MLVU, where the video-encoder-only design performs best. The reason why the video-encoder-only performs better is because it processes all 16 frames, fully capturing temporal dynamics, whereas our dual-encoder uses frame selection to select 8 key-frames plus complementary image features, slightly reducing temporal coverage. Removing frame selection (FS) in the dual-encoder achieves 48.3\% on MLVU and surpasses video-only results in. Overall, the dual-encoder design is well-suited for long videos as it can exploit complementary visual cues without sacrificing efficiency.

% Overall, these results highlight the advantage of the dual-encoder design in capturing spatiotemporal features, leading to improved video understanding.

In Tab.~\ref{tab:encoder_pooling_and_data}, we present ablation studies evaluating our dual-encoder design in Mobile-VideoGPT, alongside the contributions of our frame-selection algorithm and token projection module. Across most benchmarks, the dual-encoder consistently outperforms single-encoder variants. The exception is the MLVU benchmark, where the video-encoder-only variant achieves the highest performance. This is because the video-only model processes all 16 frames, fully capturing temporal dynamics, whereas the dual-encoder leverages frame selection to choose 8 key frames complemented by image-level features, which slightly reduces temporal coverage. Notably, when frame selection (FS) is removed in the dual-encoder, performance on MLVU increases to 48.3\%, surpassing the video-only variant, confirming that temporal coverage is critical for this benchmark. These findings provide two key insights. First, the dual-encoder design effectively captures complementary spatial and temporal cues, leading to robust performance across a wide range of tasks while maintaining efficiency. Second, for benchmarks or scenarios requiring dense temporal information, such as MLVU, there is a trade-off between efficiency and temporal completeness. 

Although the dual-encoder has slightly lower throughput (45.9 vs. 49.6 tokens/sec) than the image-only variant, it achieves around 5\% higher accuracy on challenging tasks like EgoSchema and MLVU. Our lightweight video encoder (VideoMamba with 73M parameters) adds minimal overhead, making this an efficient trade-off for real-time edge deployment with enhanced accuracy. Overall, the ablation study demonstrates that the dual-encoder design offers a favorable balance between efficiency and spatiotemporal representation, highlighting its potential for scalable, long-video understanding while maintaining generalizability across video benchmarks. To evaluate the effectiveness of our frame-selection method, we train the model without frame selection, meaning the video encoder processes all video frames. The results show that our attention-based frame-selection performs on-bar by processing all video frames while achieving higher throughput. Finally, we compare our efficient token projector against the widely used MLP-based projection, showing improved accuracy while maintaining higher throughput.

While our model demonstrates strong overall performance, its effectiveness on fine-grained tasks such as action and pose recognition is relatively limited. We attribute this to two primary design choices aimed at maximizing efficiency and enabling real-time inference: (i) using a lower spatial resolution compared to larger baselines, and (ii) constraining the number of frames to reduce temporal redundancy. Both factors, while beneficial for speed and memory efficiency, may discard subtle spatial and temporal cues that are critical for distinguishing fine-grained categories.

\begin{table}[t]
\centering
\caption{\textbf{Ablation on frame configurations and efficiency.} We compare fine-grained action and pose recognition performance against inference throughput. FS (Frame Selection) improves efficiency (Throughput) but introduces a slight trade-off in accuracy. Best results are in \textbf{bold}.}
\label{tab:finegrained_efficiency}
\setlength{\tabcolsep}{4pt}
\small
\begin{tabular}{lccc}
\toprule
Setting & Fine-grained Action $\uparrow$ & Fine-grained Pose $\uparrow$ & Throughput (fps) $\uparrow$ \\
\midrule
16 frames      & 39.8 & 32.4 & 38.5 \\
16 frames + FS & 36.5 & 28.0 & \textbf{45.9} \\
32 frames      & \textbf{43.3} & \textbf{41.6} & 30.2 \\
32 frames + FS & 41.2 & 37.8 & 36.7 \\
\bottomrule
\end{tabular}
\end{table}

To better understand this trade-off, we conducted additional ablation by varying the number of frames with our frame selection strategies. Tab.~\ref{tab:finegrained_efficiency} reports the results on fine-grained action and pose tasks. We observe that increasing the frame count from 16 to 32 improves performance because the model sees more informative temporal segments. However, this comes at the cost of higher computational overhead with lower throughput. For instance, increasing the number of input frames from 16 to 32 improves the fine-grained action by 3.7\% and fine-grained pose by 9.2\%, while reducing the throughput from 38.5 to 30.2. These findings highlight the fundamental tension between efficiency and fine-grained recognition. 
%A promising direction for future work is to incorporate adaptive mechanisms that dynamically allocate higher spatial or temporal resolution only when fine-grained distinctions are required, while maintaining efficiency for simpler cases.

\begin{figure*}[t!]
  \centering
    \includegraphics[width=0.95\linewidth]{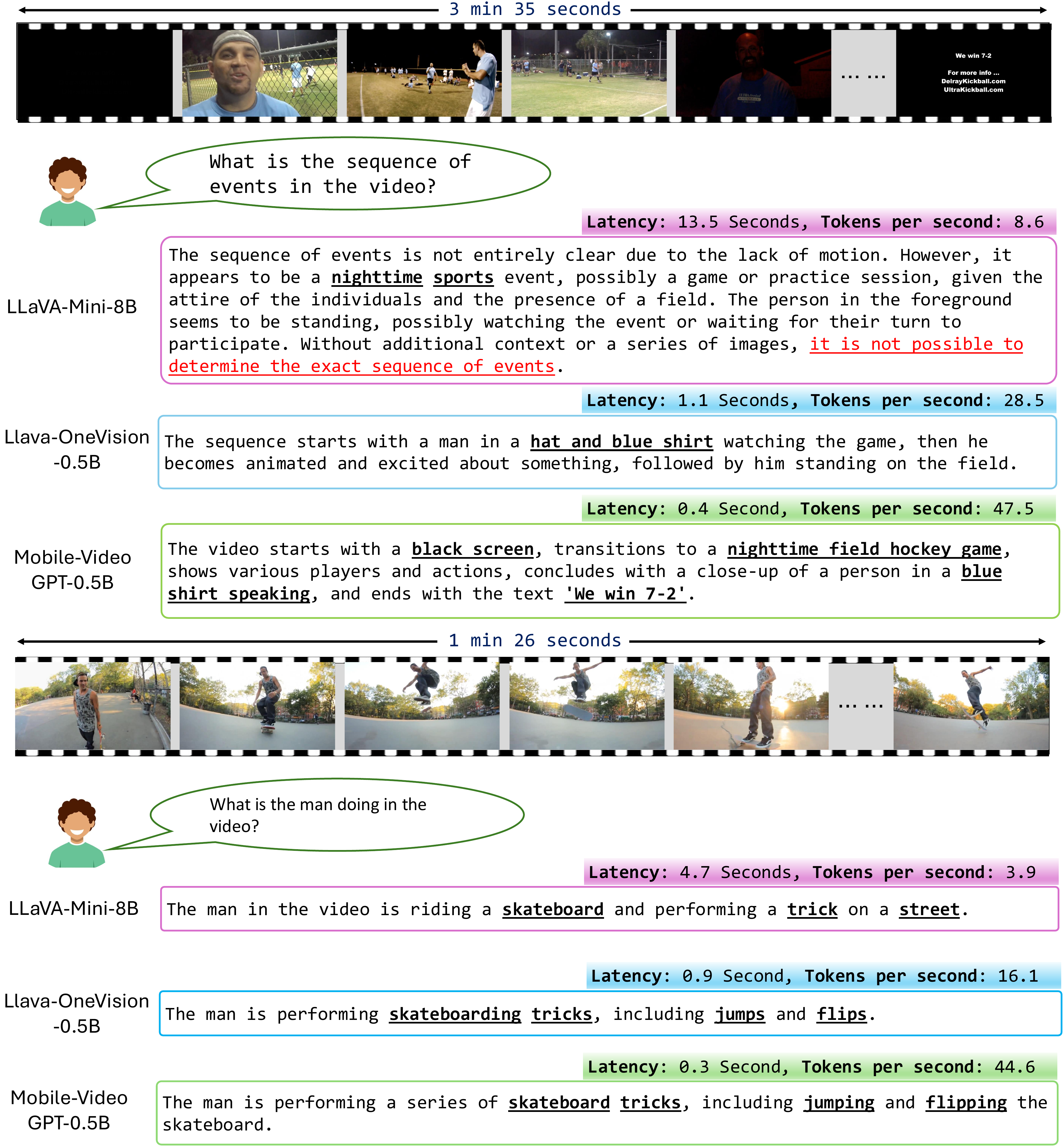}
    \caption{\textbf{Qualitative comparison of Mobile-VideoGPT-0.5B against LLaVA-OneVision-0.5B and LLaVA-Mini-8B.} Mobile-VideoGPT produces more accurate and contextually grounded responses across diverse video understanding scenarios, including temporal reasoning, action recognition, and scene comprehension. Beyond response quality, our model achieves superior efficiency, with lower latency and higher throughput (tokens per second), demonstrating that compact design need not sacrifice comprehension quality. More qualitative examples are presented in the suppl. material.}
    \label{fig:qualitative_results}
\end{figure*}

\begin{figure*}[t!]
  \centering
    \includegraphics[width=0.95\linewidth]{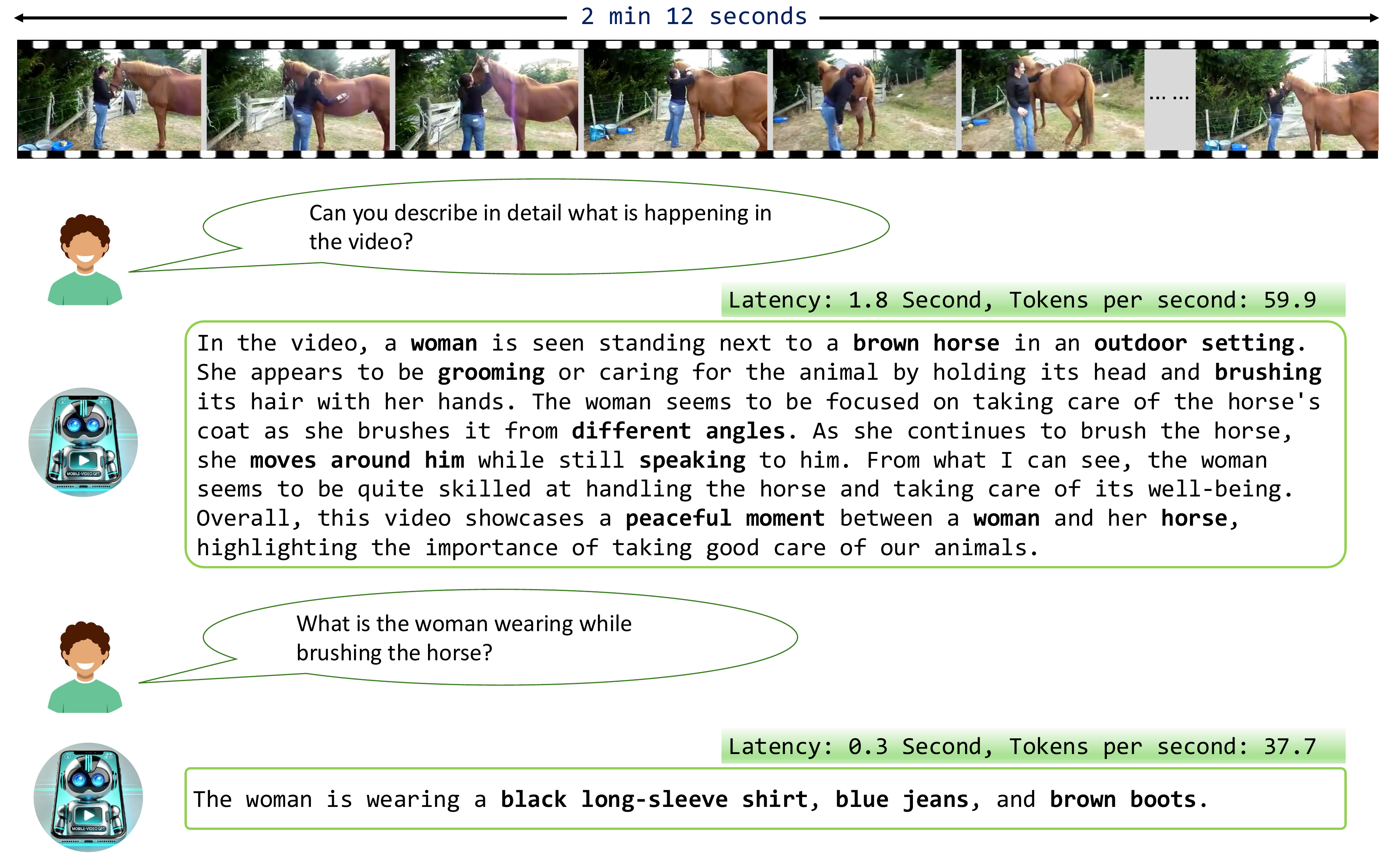}
    \caption{\textbf{Qualitative results of Mobile-VideoGPT-0.5B on open-ended video question answering.} Our model demonstrates strong video comprehension capabilities across real-world scenarios, accurately responding to both short factual and long descriptive open-ended questions. Mobile-VideoGPT produces detailed, rich, temporally coherent answers while maintaining real-time inference throughput (tokens per second), underscoring its suitability for efficient on-device video understanding.}
    \label{fig:qualitative_results_2}
\end{figure*}

\subsection{Qualitative Results}
Fig.~\ref{fig:qualitative_results} compares Mobile-VideoGPT with two recent baselines (LLaVa-Mini-8B~\cite{llavamini} and LLaVa-OneVision-0.5B~\cite{li2024llava}) on two video segments: a nighttime field hockey game (top) and a skateboarding session (bottom). The baselines provide incomplete or generic descriptions, missing key details such as the final score in the hockey game. In contrast, Mobile-VideoGPT captures the full temporal progression with detailed outputs, while also achieving over 40 tokens per second with lower latency. Mobile-VideoGPT-0.5B is 2–3$\times$ faster than LLaVa-OneVision-0.5B and 5$\times$ faster than LLaVa-Mini in the top video, and over 10$\times$ faster in the bottom video.

We show in Fig.~\ref{fig:qualitative_results_2} a representative example in which our Mobile‐VideoGPT-0.5B model responds to both detailed and concise queries regarding a short video clip. When asked to describe the scene, the model accurately explains that the woman is brushing and caring for the horse, highlighting key actions and the surrounding setting. It then handles a more focused query—what the woman is wearing—by providing a concise and correct answer, showcasing its ability to manage multi‐turn conversations of varying complexity within a single inference session. Beyond response accuracy, the model demonstrates strong temporal grounding, correctly attributing actions to the appropriate visual context without hallucination or temporal drift. Notably, the model maintains low latency (1.8 seconds for the detailed description and 0.3 seconds for the concise follow‐up) and a high token throughput, making it well-suited for real‐time, on-device video understanding tasks across a range of query types.

\section{Conclusion}
\label{sec:concl}
We tackle the computational inefficiency in video understanding through an efficient multimodal architecture designed for real-time deployment on edge devices. By combining lightweight visual encoders, efficient token projectors, and small language models, our framework reduces the computational overhead while maintaining strong performance. The innovative Attention-Based Frame-Scoring dynamically selects key-frames, while efficient token projection retains critical visual context, enabling real-time throughput. Our evaluation across six benchmarks shows that our model achieves competitive performance, while being 2$\times$ faster and 3$\times$ smaller compared to Llava-One-Vision-0.5B on Jetson Orin Nano, which highlights the model's strong practicality on edge devices. 
%Our model is compact, with a size of 1GB, and requires 3GB VRAM. 

% \section{Acknowledgment}
% \label{sec:Acknowledgment}
% The computations were enabled by resources provided by NAISS at Alvis
% partially funded by Swedish Research Council through grant agreement no. 2022-06725, LUMI
% hosted by CSC (Finland) and LUMI consortium, and by Berzelius resource provided by the Knut and
% Alice Wallenberg Foundation at the NSC.
%This efficiency narrows the gap between high-performance video understanding and real-world deployment on edge devices.

{
    \small
    \bibliographystyle{splncs04}
    \bibliography{main}
}
\clearpage

\section{Supplementary Material}
In this section, we provide more details regarding the following:-

\begin{itemize}

\item Discussion about Fine-Grained Tasks vs. Efficiency
\item Additional Benchmark Details
\item Additional Training Details
\item Comparison between the visual tokens for Mobile-VideoGPT and LLaVa-OneVision 
\item Additional Qualitative Results
\end{itemize}

\subsection{Discussion about Fine-Grained Tasks vs. Efficiency}
As discussed in the main paper, the limited fine-grained recognition of \ours stems from two 
efficiency-driven design choices: lower spatial resolution and a constrained frame count. 
Tab.~\ref{tab:finegrained_efficiency} further examines this trade-off by varying the number 
of frames alongside our frame selection (FS) strategy. Increasing the frame count from 16 to 
32 yields consistent gains in fine-grained action (+3.7\%) and pose (+9.2\%) recognition by 
capturing more informative temporal segments, albeit at the cost of reduced throughput 
(38.5$\rightarrow$30.2~fps). Applying FS partially recovers this efficiency gap (\eg, 32 
frames + FS reaches 36.7~fps) with only a modest accuracy drop, highlighting its role as an 
effective efficiency--accuracy balancer. 

These findings underscore the fundamental tension 
between efficiency and fine-grained recognition. A promising direction for future work is to 
incorporate adaptive mechanisms that dynamically allocate higher spatial or temporal resolution 
only when fine-grained distinctions are required, while maintaining efficiency for simpler cases.

\subsection{Additional benchmark details}
\noindent

\textbf{ActivityNet-QA}~\cite{yu2019activitynet} is a short question-answering dataset derived from the popular ActivityNet~\cite{yu2019activitynet} dataset. The test set consists of 8K samples. Following Video-ChatGPT~\cite{Maaz2023VideoChatGPT}, we employ GPT-3.5-Turbo for evaluation. 

\textbf{EgoSchema}~\cite{mangalam2023egoschema} is an MCQ-based benchmark where the videos are sourced from Ego4D~\cite{grauman2022ego4d}. The test set consists of 5K samples, and the results are compiled by submitting the model's predictions to the online evaluation server. 

\textbf{MLVU}~\cite{zhou2024mlvu} is a multimodal long video understanding benchmark encompassing nine distinct tasks, including MCQs as well as open-ended generation tasks. Following LLaVA-OneVision~\cite{li2024llava}, we report results on the dev set, which consists of 2,593 samples in total. Although our model is not specifically designed for long video understanding, it achieves competitive performance.

\textbf{MVBench}~\cite{li2023mvbench} is a comprehensive multimodal video understanding benchmark containing 20 sub-tasks, including Action Sequence (AS), Action Prediction (AP), Action Antonym (AA), Fine-grained Action (FA), Unexpected Action (UA), Object Existence (OE), Object Interaction (OI), Object Shuffle (OS), Moving Direction (MD), Action Localization (AL), Scene Transition (ST), Action Count (AC), Moving Count (MC), Moving Attribute (MA), State Change (SC), Fine-grained Pose (FP), Character Order (CO), Egocentric Navigation (EN), Episodic Reasoning (ER), and Counterfactual Inference (CI). Each sub-task includes 200 MCQs, totaling 4K MCQs in the benchmark. 

\textbf{NExTQA}~\cite{xiao2021next} is a video question-answering benchmark that focuses on explaining temporal actions in videos. Following previous approaches, we evaluate on the MCQ test set consisting of 8,564 samples, targeting causal action reasoning, temporal action reasoning, and common scene comprehension. 

\textbf{PerceptionTest}~\cite{puatruaucean2023perception} is designed to assess the perception and reasoning capabilities of multimodal models. Following previous approaches, we report results on the validation set, which consists of 19,140 samples.

\begin{table*}[t]
\centering
\caption{\textbf{Training configuration of \ours across three stages.} Stage 1 pre-trains the image projection layer while keeping all other modules frozen. Stage 2 extends pre-training to the video projection layer. Stage 3 performs full instruction tuning with the language model unfrozen. All stages use AdamW with cosine decay scheduling.}
\small
\setlength{\tabcolsep}{10pt}
\renewcommand{\arraystretch}{1.1}
\resizebox{\textwidth}{!}{%
\begin{tabular}{llccc}
\toprule
\multicolumn{2}{l}{\multirow{2}{*}{\textbf{Settings}}}
& \textbf{Stage 1}
& \textbf{Stage 2}
& \textbf{Stage 3} \\
\multicolumn{2}{l}{}
& \textit{Image Proj. Pre-training}
& \textit{Video Proj. Pre-training}
& \textit{Instruction Tuning} \\
\midrule
\multirow{4}{*}{\textbf{Modules}}
& Image Encoder        & Frozen    & Frozen    & Frozen    \\
& Video Encoder        & Frozen    & Frozen    & Frozen    \\
& Small Language Model & Frozen    & Frozen    & Trainable \\
& Projection           & Trainable & Trainable & Trainable \\
\midrule
\multirow{6}{*}{\textbf{Hyperparameters}}
& Batch Size      & 128          & 128          & 64           \\
& Learning Rate   & 1e-3         & 1e-3         & 2e-4         \\
& Schedule        & Cosine decay & Cosine decay & Cosine decay \\
& Warmup Ratio    & 0.03         & 0.03         & 0.03         \\
& Optimizer       & AdamW        & AdamW        & AdamW        \\
& Epochs          & 2            & 2            & 2            \\
\bottomrule
\end{tabular}}
\label{tab:training}
\end{table*}

\subsection{Additional training details}
\noindent
We adopt a three‐stage training pipeline as summarized in Tab.~\ref{tab:training}. In \textbf{Stage 1 (Image Projector Pre‐training)}, the image and video encoders, as well as the small language model, are kept frozen. Only the efficient image token projection module is trainable. We use a batch size of 128, a learning rate of $1\times10^{-3}$, and a cosine decay schedule with a warmup ratio of 0.03 for two epochs, optimized by AdamW.  \textbf{Stage 2 (Video Projector Pre‐training)} follows the same hyperparameter settings, again freezing the image/video encoders and the small language model, and training only the projection module. \textbf{Stage 3 (Instruction Tuning)} keeps the encoders frozen but makes both the projection module and the small language model trainable. The latter is trained with LoRA, configured as $lora_{r}=128$ and $lora_{\alpha}=256$. In this stage, we use a smaller learning rate of $2\times10^{-4}$, a batch size of 64, and maintain the same optimizer and schedule strategy for two additional epochs. Pre‐training of the image and video projectors takes approximately 16 hours on 8~A100 40\,GB GPUs, while the instruction fine‐tuning stage requires around 36 hours.

\begin{table*}[t]
\centering
\caption{\textbf{Token efficiency comparison between LLaVA-OneVision and \ours.} \ours uses efficient token projection (ET$_\text{Proj}$) combined with frame sampling~(FS), reducing total visual tokens from $6{,}272$ to $2{,}696$, a \textbf{2.3$\times$ reduction}, while separately handling image and video tokens through dedicated projection heads.}
\small
\setlength{\tabcolsep}{8pt}
\renewcommand{\arraystretch}{1.1}
\resizebox{\textwidth}{!}{%
\begin{tabular}{llccc}
\toprule
\textbf{Model} & \textbf{Projection Strategy} & \textbf{Image Tokens} & \textbf{Video Tokens} & \textbf{Total Tokens} \\
\midrule
LLaVA-OneVision      & MLP$_\text{Proj}$             & $32 \times 196 = 6{,}272$ & --                        & $6{,}272$ \\
\ours (Ours)         & ET$_\text{Proj}$ + FS         & $16 \times 144 = 2{,}304$ & $8 \times 49 = 392$       & $2{,}696$ \\
\bottomrule
\end{tabular}}

\label{tab:model-projection-strategy}
\end{table*}

\subsection{Comparison between the visual tokens for Mobile-VideoGPT and LLaVa-OneVision }
Tab.~\ref{tab:model-projection-strategy} compares Mobile‐VideoGPT against LLaVA‐OneVision in terms of projection strategies and their resulting token counts. LLaVA‐OneVision uses a simple MLP‐based projector (\textsc{MLP}\textsubscript{Proj}) that handles 32 input frames, each producing 196 tokens, for a total of 6272 tokens. By contrast, our method relies on an Efficient Token Projection layer (\textsc{ET}\textsubscript{Proj}) combined with frame selection (\textsc{FS}). For images, we consider 16 frames, each represented by 144 tokens, yielding 2304 tokens. For video, we select 8 of those frames, each producing 49 tokens, resulting an additional 392 tokens. Altogether, we generate a total of 2696 visual tokens, substantially reducing the visual token load while simultaneously extending the model’s capabilities to video data, thereby underscoring the efficiency and flexibility of Mobile‐VideoGPT.

\subsection{Additional qualitative results}

\begin{figure*}[t!]
  \centering
    \includegraphics[width=1.0\linewidth]{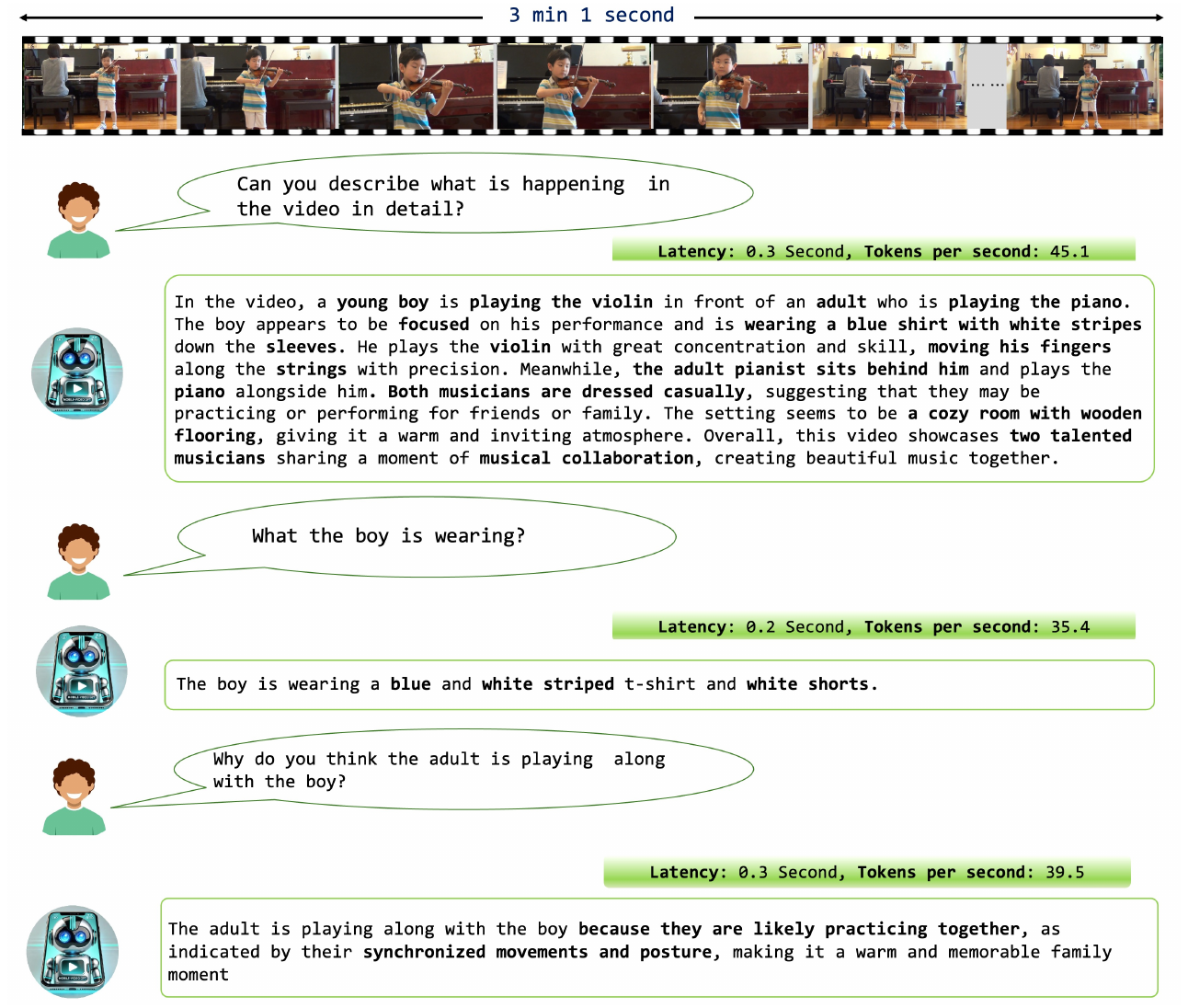}
    \caption{\textbf{Qualitative results of Mobile-VideoGPT-0.5B on open-ended video question answering.} Our model demonstrates strong video comprehension capabilities across real-world scenarios, accurately responding to both short factual and long descriptive open-ended questions. Mobile-VideoGPT produces detailed, rich, temporally coherent answers while maintaining real-time inference throughput (tokens per second), underscoring its suitability for efficient on-device video understanding.}
    \label{fig:suppl_material}
\end{figure*}

We show in Fig~\ref{fig:suppl_material} additional representative example in which our \ours 0.5B model 
responds to both detailed and concise queries regarding a 3-minute video clip. When asked to 
describe the scene, the model accurately identifies a young boy playing the violin accompanied 
by an adult pianist, correctly noting details such as the boy's attire, their synchronized 
performance, and the warm setting. It then handles more focused follow-up queries---what the 
boy is wearing and why the adult is playing along---providing concise and correct answers, 
showcasing its ability to handle multi-turn conversations of varying complexity. Notably, the 
model maintains consistently low latency (0.2--0.3 seconds per response) and high token 
throughput (35--45 tokens per second), making it practical for real-time video understanding 
tasks on resource-constrained devices.

\end{document}